\documentclass[wcp]{jmlr}

\usepackage{longtable}

\usepackage{booktabs}
\usepackage{natbib}
\usepackage{soul}
\usepackage{url}
\usepackage{caption}
\usepackage{mathtools}
\usepackage{booktabs}
\usepackage{multirow}
\usepackage{setspace}
\usepackage[noend]{algorithm2e}
\usepackage{array}
\usepackage{threeparttable}
\usepackage{bbm}
\usepackage{makecell}
\makeatletter
\let\Ginclude@graphics\@org@Ginclude@graphics 
\makeatother

\jmlrvolume{1}
\jmlryear{2022}

\title{Efficient Deep Clustering of Human Activities and\\ How to Improve Evaluation}
\author{Louis Mahon and Thomas Lukasiewicz}

\begin{document}

\maketitle

\begin{abstract}
There has been much recent research on human activity re\-cog\-ni\-tion (HAR), due to the proliferation of wearable sensors in watches and phones, and the advances of deep learning methods, which avoid the need to manually extract features from raw sensor signals. A significant disadvantage of deep learning applied to HAR is the need for  manually labelled training data, which is especially difficult to obtain for HAR datasets. Progress is starting to be made in the unsupervised setting, in the form of deep HAR clustering models, which can assign labels to data without having been given any labels to train on, but there are problems with evaluating deep HAR clustering models, which makes assessing the field and devising new methods difficult. In this paper, we highlight several distinct problems with how deep HAR clustering models are evaluated, describing these problems in detail and conducting careful experiments to explicate the effect that they can have on results. We then discuss solutions to these problems, and suggest standard evaluation settings for future deep HAR clustering models. Additionally, we present a new deep clustering model for HAR. When tested under our proposed settings, our model performs better than (or on par with) existing models, while also being more efficient and better able to scale to more complex datasets by avoiding the need for an autoencoder.    
\end{abstract}

\section{Introduction}
Human activity recognition (HAR), the task of automatically determining the activity that a person is performing based on recorded data, has a number of important applications. It is of interest to healthcare research, as it can provide a direct measure of exercise frequency and intensity. The World Health Organization lists inactivity as the fourth leading risk factor for mortality, and estimates that over 30\% of adults are insufficiently active.\footnote{https://www.who.int/data/gho/indicator-metadata-registry/imr-details/3416} However, such estimations are difficult. Self-report does not give a reliable measure of exercise, as patients tend to significantly over-report \citep{mcconnell2018mobile}, so being able to directly monitor human activity is desirable. HAR is also used in wearable sports technology. Sports watches, for example, provide users with a breakdown of how much time they spend sitting, standing, and walking. Globally, the wearable technology market was valued at \$41bn in 2019, and it is forecasted to grow to \$114bn by 2028.\footnote{https://www.grandviewresearch.com/industry-analysis/wearable-technology-market} 

The recorded data on which human activity recognition is based can come from three different types of device: video recorders, ambient sensors, and wearable sensors. 
This data is then input to a recognition model, to infer the activity being performed. If video recorders are used, then the task is one of computer vision, if ambient or wearable sensors are used, then the task is a form of signal processing. The difference between ambient and wearable sensors is that the former stays at a fixed location in the environment, and the latter is attached to the human performing the activity. In this paper, we focus primarily on HAR from wearable sensors. There are two types of wearable sensors, accelerometers, which measure acceleration in three spatial dimensions, and gyroscopes, which measure orientation and angular momentum. 

Raw sensor data cannot always be easily interpreted by human inspection, which has two important consequences. Firstly, it can make feature engineering difficult. For example, in the case of gyroscope readings, we do not know, a priori, what the relevant differences are between the signals for certain activities, especially those that are similar, such as walking upstairs vs.\ downstairs. While engineered features have been used with some success (see Section \ref{sec:related-work}), these are mostly statistical features, rather than features that leverage domain knowledge. Deep learning, which can learn to automatically extract features, is therefore an attractive approach to HAR. The second important consequence is that HAR data are very difficult to label. Labelled data are always more expensive and time-consuming to obtain than unlabelled data, but this is especially the case for sensor-based HAR data, because humans cannot provide annotations just by looking at the sensor readings. Instead, annotators must directly observe a subject, which requires taking them into the lab, or be given a video of the performed actions, which requires subjects to remember to film themselves while using the sensors outside the lab. There is therefore a need for models that can operate without labelled data, as has been noted in two recent survey papers \citep{wang2019deep,chen2021deep}. This is one reason HAR clustering is of value. If it was solved very accurately, so that all instances of the same activity were clustered together and all instances of different activities were clustered separately, then, once the cluster labels had been aligned with activity names such as `walking', the HAR classification problem would also have been solved, and its solution would not have required any labelled data at all. Another advantage of HAR clustering is that, even in the absence of a very accurate solution, it can shed light on the most appropriate set of classes into which activities should be partitioned. 
For example, some datasets distinguish between `walking' and `fast walking', while some others just use a single class `walking'; similarly for `running' and `jogging'. If clustering shows there to be a significant difference between walking and fast walking, this is evidence that such a distinction is warranted. This use is not explored further here, though it has been in previous works \citep{mejia2017evaluate}.

For these reasons, HAR clustering has received significant research attention. This paper focuses on deep HAR clustering, i.e., clustering HAR data using a deep neural network for feature extraction. Recent years have seen some works applying deep learning to HAR clustering \citep{mcconville2021n2d,sheng2020unsupervised,ma2021unsupervised}. However, progress has been obstructed in deep HAR clustering, and HAR clustering more generally, by a lack of agreed evaluation standards. Different works test on different datasets, many of them private, and some crucial details are left out when describing the exact evaluation settings. In particular, the distinction between subject-dependent and subject-independent clustering is often not made explicit, even though it greatly affects the results. 

As well as highlighting these problems and describing more rigorous evaluation settings that can address them, we propose a new deep HAR clustering model and test it under these settings, showing that it outperforms existing methods (insofar as a comparison can be made, with respect to the above points). We then present ablation studies on its main components. 
Ours is the first deep HAR clustering model not to require the reconstruction loss of an autoencoder, making it more efficient and better able to scale to more complex datasets. 
Below is a brief summary of our contributions. 
\begin{itemize}
    \item We explicate differences in evaluation procedures for deep HAR clustering, and show empirically that these differences can affect performance metrics. While our main focus is on deep HAR clustering, much of our analysis, including the important distinction between subject-dependence and subject-independence, applies to HAR clustering in general.
    \item We discuss suitable evaluation settings to use for HAR clustering. Adoption of our recommendations by future works will enable a direct comparison and benchmarking of deep HAR clustering models (and HAR clustering models more generally), and thus accelerate progress in the field. 
    \item We describe a streamlined and scalable deep HAR clustering model. 
    On six public datasets, this model performs better than or on par with existing models (insofar as they can be compared). We also present ablation studies showing the contributions of its components. 
\end{itemize}
The rest of this paper is organized as follows. Section \ref{sec:related-work} provides an overview of related work. Section \ref{sec:problems-with-existing} explicates the short\-comings of existing evaluation methods for HAR clustering. In Section \ref{sec:method}, we describe our new method for HAR. Section \ref{sec:results} then presents the results of our method under our proposed evaluation settings, and Section~\ref{sec:conlusion} summarizes our work.
Note that the code of this paper will be released on publication.

\section{Related Work} \label{sec:related-work}

Machine learning has been identified as a promising approach to HAR since at least 2000 \citep{hongeng2000representation}, with early works using, e.g., naive Bayes \citep{tapia2004activity}, 
support vector machines \citep{he2009activity}, and generalized discriminant analysis \citep{khan2010human}. These machine learning algorithms require feature engineering, and in the case of HAR, this is commonly done by taking statistical quantities such as mean and higher moments of the raw signal, in both the time and frequency domains. These can be combined with more bespoke features \citep{zhang2011feature,he2009activity}.


Deep learning is a form of machine learning that does not require hand-crafted features, but rather can learn to extract features from a raw input.
Applied to HAR, not only does deep learning avoid the need for feature engineering, but has also been shown to achieve better accuracy than feature-engineered models \citep{alsheikh2016deep,ferrari2019hand}. Network architectures include convolutional neural networks (CNNs)
\citep{yang2015deep,chen2015deep,jiang2015human,ronao2015deep} and recurrent neural networks (RNNs) 
\citep{inoue2018deep,singh2017human}. See \citep{hammerla2016deep} for an empirical comparison of CNNs and RNNs for HAR. See \citep{chen2021deep,wang2019deep} for summaries of recent deep HAR models. 

There has been increasing interest in reducing the supervision needed for HAR. The first efforts in this direction were semi-supervised methods trained on unlabelled data alongside labelled data \citep{li2011incorporating}, that used unlabelled data for pretraining \citep{li2014unsupervised,alsheikh2016deep}, or that evaluatinvestigated the optimality of the label-set by comparing to cluster-labels \citep{mejia2017evaluate}; bHowever, training these semi-supervised models still requires some labelled data.
According to a recent survey on HAR, the need for fully unsupervised models is urgent due to the difficulty obtaining labels \citep{wang2019deep}. Another \citep{chen2021deep} discussed the advantages of unsupervised HAR models but notes a disadvantage that most are restricted to feature extraction and cannot produce labels. Deep clustering models can redress this problem by interpreting cluster membership as labels. 

As is the case for supervised models, most HAR clustering models begin with a feature extraction stage. While some works apply clustering algorithms directly to the raw sensor signal \citep{trabelsi2013unsupervised}, the most common approach is to first extract features from the sensor signal, and then cluster the extracted features. Initial HAR clustering models performed feature extraction by computing statistical quantities \citep{kwon2014unsupervised,machado2015human,lu2017towards}. In \citep{he2017unsupervised}, statistical features are replaced by the discrete wavelet packet transform 
followed by principal components analysis for dimensionality reduction. Clustering is performed using fuzzy c-means with a novel initialization method based on cosine similarity. Another HAR clustering model is proposed in \citep{sheng2020unsupervised}, which includes a deep autoencoder in the feature extraction stage with two additional loss terms to encourage locality and temporal consistency. Statistical feature extraction is dispensed with completely in \citep{ma2021unsupervised}, replaced with a CNN-BiLSTM autoencoder plus pseudo-label training \citep{caron2018deep}. In \citep{mcconville2021n2d}, a deep autoencoder is combined with UMAP \citep{mcinnes2018umap}, for dimensionality reduction, followed by clustering using a Gaussian mixture model (GMM).

\section{Problems with Existing Literature} \label{sec:problems-with-existing}
We identify three problems with the existing field of HAR clustering, shown with respect to existing works in Table \ref{tab:previous-works}.
\begin{itemize}
    \item Different works often report on different datasets using different metrics. Many works report results on their own new dataset, and so cannot compare to prior works. Additionally, the datasets are often private.
    \item The exact evaluation criteria are unclear. There are multiple ways of evaluating the performance of a model, which can give significantly different results. Of particular importance is whether each subjects' data are clustered individually or whether all data are clustered together.  
    \item Code is not released, making reproducibility difficult or impossible.
\end{itemize}

These problems make it hard for new researchers in the field to assess the best existing models from which to build, and difficult for them to determine whether they have improved over these existing models. Consequently, progress is held back. 

In the following sections, we address these issues. Section \ref{subsec:datasets} describes our choices of datasets on which to evaluate performance, and the reasoning behind these choices. Section \ref{subsec:ambiguous-eval} discusses the effect of different evaluation settings on perfor\-mance metrics, and demonstrates these effects empirically, by showing that the same model can produce significantly different results in different evaluation settings.

\subsection{Datasets} \label{subsec:datasets}

We select six suitable wearable-sensor HAR datasets for measuring clustering performance: Physical Activity Monitoring (PAMAP2) \citep{reiss2012introducing}, Human Activity Recognition using Smartphones (UCI-Sm) \citep{anguita2013public}, WISDM-v1 \citep{kwapisz2011activity}, WISDM-watch \citep{weiss2019smartphone}, Realistic Sensor Displacement (REALDISP) (under the `ideal placement' setting) \citep{banos2012benchmark}, and Heterogeneous Human Activity Recognition \citep{stisen2015smart}. The details of subjects, activities, and sensors for each dataset are shown in Table \ref{tab:dataset-info}. There are three reasons for selecting these particular datasets:
\begin{itemize}
    \item They are all easily accessible in the UCI repository.
    \item They have been used by some previous works, and so enable comparison. Although, as shown in Table \ref{tab:previous-works}, there is not as much consistency in the use of datasets as would be ideal. 
    \item They vary in number of activities, number of data points, number of subjects, and number of sensor channels, helping to measure generalization ability. UCI-Sm and WISDM-v1 are smaller datasets, with only a few channels and activity classes.
    The other four datasets are more complex. WISDM-watch is unique in having a large number of users, 51, and HHAR is unique in having a large number of data points. REALDISP is a large dataset with many sensors channels. This set of datasets thus tests a model's performance in a range of settings, from a small to a large number of users, from a small to a large number of clusters and from simple hardware to many wearable sensors with a rich array sensor channels. 
\end{itemize}

\begin{table*}[h] 
    \centering
    \caption{Previous HAR clustering models relative to the evaluation criteria outlined in Section \ref{sec:problems-with-existing}. There are a number of different datasets and metrics, but none releases their code, and almost all are ambiguous as to subject independence (indicated S-dep below.)\vspace*{-0ex}}
     \label{tab:previous-works}
   
    \begin{threeparttable}
       \resizebox{1.0\textwidth}{!}{ \begin{tabular}{|c|c|c|c|c|}
         \hline
         Name                             & Datasets          &   Metrics                             &   \makecell{Code \\ Released}  & S-Dep \\
         \hline \hline
         \cite{kwon2014unsupervised}      &   own (private)   &   ACC, NMI                        &   no & unlcear \tnote{\dag} \\
         \hline
         \cite{trabelsi2013unsupervised}  &   own (private)   &   ACC, precision, recall          &   no & unclear\\
         \hline
         \cite{lu2017towards}             &   own (private)   &   \makecell{ACC, precision, \\ recall, specificity, \\ ARI, FM-index} & no & unclear \\
         \hline
         \cite{machado2015human}          &   own (private)   &   ACC                             &   no & both \\
         \hline
         \cite{he2017unsupervised}        &   WISDM-v1        &   RI, ARI         &   no & unclear\tnote{\ddag} \\
         \hline
         \cite{sheng2020unsupervised}     &   PAMAP2, SBHAR, REALDISP    &   ACC, ARI, NMI        &   no & unclear  \\
         \hline
         \cite{he2018wavelet}             &   DSAD            &   RI, ARI, FM-index               &   no & yes \\
         \hline
         \cite{ma2021unsupervised}      &  \makecell{HAR, MotionSense\tnote{1},  \\ MobiAct\tnote{2}, own (private)}   &   precision, recall, F1, NMI 
         & no  &   unclear \\
         \hline
         \makecell{\textcolor{blue}{Dobbins and} \\ \textcolor{blue}{Rawassizadeh (2018)}}&  HHAR   &   silhouette-index                            &   no & unclear \tnote{*}\\
         \hline
         ours      &  \makecell{PAMAP2, UCI-Sm, \\ WISDM-v1, WISDM-watch, \\ REALDISP, HHAR} &   ACC,ARI,NMI,F1        &   \makecell{yes (on \\ publication)} & yes\\
         \hline
    \end{tabular}}
    \end{threeparttable}
    \begin{tablenotes}
        \footnotesize
        \item[1]{$^1$\cite{altun2010comparative}}
        \item$^2${\cite{malekzadeh2018protecting}}
        \item$^\dag$ mentions different cluster sizes for different subjects, implying subject-dependence
        \item$^\ddag$displays confusion matrix referring to one subject only
        \item$^*$ discusses subject heterogeneity within activity classes, implying subject-independence
    \end{tablenotes}
\end{table*}

\begin{table}
   \medskip 
  \centering
  \footnotesize
  \caption{Evaluation  under the four settings corresponding to the two ambiguities described in Section \ref{subsec:ambiguous-eval}. Window-wi\-se vs.\ point-wise does not affect results, but subject-dependent vs.\ subject-independent does. The subject-dependent settings performs substantially better across all datasets and metrics. It is therefore essential that HAR clustering works specify whether they are testing in the subject-dependent or subject-independent setting.\vspace*{-0ex}} \label{tab:baseline-results}
  \resizebox{1.0\textwidth}{!}{\begin{tabular}{|*{6}{c|}}
    \hline
    &  & \makecell{window-wise \\ subject-dependent} & \makecell{point-wise \\subject-dependent} & \makecell{window-wise\\ subject-independent}  & \makecell{point-wise\\ subject-independent} \\
         \hline\hline
    
    \multirow{4}{6em}{PAMAP}        & ACC & 66.28 & 66.27 & 48.30 & 47.35 \\ 
                                    & NMI & 64.95 & 64.80 & 46.61 & 48.14 \\
                                    & ARI & 50.57 & 50.54 & 30.44 & 31.31 \\
                                    & F1 & 65.63 & 65.57 & 45.26 & 45.73 \\
                            \hline
    \multirow{4}{6em}{UCI-Sm}       & ACC & 50.57 & 50.73 & 38.95 & 38.93 \\ 
                                    & NMI & 56.36 & 56.77 & 35.59 & 35.57 \\
                                    & ARI & 39.57 & 39.50 & 23.94 & 23.92 \\
                                    & F1 & 46.74 & 46.88 & 35.32 & 35.28 \\
                             \hline
    \multirow{4}{6em}{WISDM-v1}     & ACC & 72.15 & 72.33 & 50.14 & 50.04 \\ 
                                    & NMI & 69.44 & 69.40 & 38.78 & 38.80 \\
                                    & ARI & 60.05 & 60.03 & 33.07 & 33.06 \\
                                    & F1 & 64.88 & 64.95 & 38.91 & 38.88 \\
                             \hline
    \multirow{4}{6em}{WISDM-watch}     & ACC & 78.40 & 78.48 & 25.58 & 25.56 \\ 
                                    & NMI & 84.68 & 84.71 & 28.38 & 28.36 \\
                                    & ARI & 72.22 & 72.17 & 12.6 & 12.59 \\
                                    & F1 & 77.61 & 77.67 & 25.32 & 25.31 \\
                             \hline
    \multirow{4}{6em}{REALDISP}     & ACC & 89.60 & 89.60 & 51.37 & 51.36 \\ 
                                    & NMI & 93.87 & 93.97 & 71.80 & 71.79 \\
                                    & ARI & 88.60 & 88.53 & 43.91 & 43.87 \\
                                    & F1 & 84.36 & 84.38 & 47.00 & 46.99 \\
                             \hline
    \multirow{4}{6em}{HHAR}         & ACC & 53.80    & 53.80  & 47.93  & 48.25 \\
                                    & NMI & 51.62    & 51.62  & 38.42  & 38.52 \\
                                    & ARI & 38.14    & 38.07  & 25.60  & 25.78 \\
                                    & F1 & 52.57    & 52.48  & 49.08  & 49.36 \\
   \hline
  \end{tabular}}
  
\end{table}

\begin{table*}[] 
    \caption{Information on each of the datasets on which we report results. Accel = 3d accelerometer, gyro = 3d gyroscope, and magneto = 3d magnetometer. Total time points = the sum of time points across all users, after discarding those without labels and those with missing data.\vspace*{-0ex}}
    \label{tab:dataset-info}
    \centering
    \resizebox{\textwidth}{!}{
        \begin{tabular}{|c|c|c|c|c|c|c|}
         \hline
         Name         & \makecell{Date \\ Released} &   \makecell{Number of \\ Activities}  &   \makecell{Number of \\ subjects}   & Sensors   & Channels & \makecell{Total \\ Time Points}\\
         \hline\hline
         PAMAP2       &   2012        &   12      &   9       & \makecell{3 x (accel, \\ gyro, magneto)}        & 51    & 1921431 \\
         UCI-Sm       &   2013        &   6       &   30      & phone accel and gyro                  & 6     & 71968 \\
         WISDM-v1     &   2011        &   5       &   36      & phone accel                           & 3     & 1085363 \\
         WISDM-watch     &   2019        &   18      &   51      & \makecell{phone and watch \\ accel and gyro}        & 12    & 3635842 \\
         REALDISP     &   2012        &   33      &   17      & \makecell{9 x (accel,gyro, \\ magneto, orient)}    & 117   & 669618 \\
         HHAR         &   2015        &   6      &   9        & 2 x (accel, gyro)                     & 12    & 11279265 \\
         \hline
    \end{tabular}
    }
    \end{table*}

\subsection{Ambiguous Evaluation Settings} \label{subsec:ambiguous-eval}
We identify two ambiguities in how HAR clustering models are evaluated, subject-dependent vs.\ subject-independent, and window-wise vs.\ point-wise. The former refers to whether clustering was performed on all subjects' data at once (subject-independent) or on each subjects' data separately (subject-dependent). For supervised models, specification of the train-test split can disambiguate subject-dependence vs.\ independence, by specifying that, e.g., data for users $X,Y,Z$ was used for testing. Clustering, however, does not typically use a train-test split, and the question of subject-dependence vs.\ independence is almost always unclear (see Table \ref{tab:previous-works}). The latter refers to whether data points are taken to be the sliding window or each time point. Each time point has one label, but training collates multiple time points into windows. (The window-size is $512$ in all our experiments.) There is ambiguity as to whether data points should be taken to be the windows or the time points. This type of ambiguity can be present in supervised models too.

We empirically investigate the effect of these two factors, by training and testing a simplified version of our proposed model (described in Section \ref{sec:method}) under the four resulting settings.  The results are displayed in Table \ref{tab:baseline-results}. 

Columns one and two are subject-dependent. They train a separate clustering model on each subject and average the results across these separate models, weighting each subject by their number of data points. Columns three and four are subject-independent. They train a single clustering model and cluster all subjects' data at once. The subject-dependent models outperform the subject-independent models by a large margin. This is in keeping with results from the supervised domain, where it has been noted that training on some data from the test subject significantly improves performance \citep{reiss2012introducing,suh2021adversarial}, which suggests the existence of subject-specific features in how activities are performed.  The large difference in results between these two settings, evidenced in \mbox{Table \ref{tab:baseline-results}}, means that it is crucial that models specify which they are using. Moreover, the two are fundamentally different tasks, one discovering patterns in the activity signal of a specific user, and the other learning generalized activities, independent of who is performing them. 

Columns one and three in Table \ref{tab:baseline-results} are window-wise, while columns two and four are point-wise. The former treats each window as a data point whose label is the most commonly occurring label across all time points in that window. The latter treats each time point as a data point. Using overlapping windows means that each time point appears in multiple windows. In order to produce a single predicted label for each time point, the window-wise setting takes the most commonly occurring label across the multiple windows that contain that time point. (We also explored other means of combining these multiple labels, with very similar results.) There is essentially no difference between window-wise and point-wise evaluation so, although existing works are ambiguous as to which is being employed, this ambiguity does not prevent a clear assessment of performance. The results presented in Section \ref{sec:results} are all in the point-wise setting.

\section{Our Method} \label{sec:method}
 
\begin{algorithm}[t]
\caption {Training algorithm} \label{alg:method}
$X \gets$ data;

$Mask,SemiMask \gets$ $X$;

$Final \gets$ empty hash table;
\While{$|Final|$ increases} {
    \For{$i=1,\dots, 10$} {
         initialize encoder, encode $X$ and UMAP to $\mathbb{R}^{2}$
         cluster using HMM, giving labels $c(x)$ and probabilities $p(x)$
        \For{$x \in X$} {
            \eIf{$p(x)<.95$ or ($i > 1$ and $c(x) \neq c'(x)$)} {
                 remove $x$ from $Mask$ and $SemiMask$
                }
            { add $x$ to $SemiMask$}
        }
        \For{$epoch=1,\dots, 5$} {
            \For {$x \in X$} {
                \uIf{$x \in Mask$}{
                     train on $x,c(x)$
                    }
                \uElseIf{$x \in SemiMask$} {
                     train on $x,c(x)$, weighted by $0.5$
                }
            }
         $c'(x) \gets c(x)$
        }
    }
    \For {$x \in Mask$} {
         $Final[x] \gets c(x)$
    }
}
\For{$x \in X\setminus Final$} {
         $Final[x] \gets c(x)$
}
\end{algorithm}

Our model is based on the technique of pseudo-label training \citep{caron2018deep}, extended with a novel method for selecting points on which to pseudo-label train. Pseudo-label training clusters the output of an encoder, then uses cluster labels as classification targets, and trains on these targets to refine the weights of the encoder. It allows the encoder weights to be interatively refined. However, the pseudo-labels are noisy and often incorrect. Previous works \citep{mahon2021selective,mrabah2020deep} have shown that filtering out the least confident labels can improve performance. We propose a novel method of filtering these labels. At each iteration, we train only on those data points that received the same cluster label as they did in the previous iteration.
We also implement a graded label filtering, by double-weighting the training updates for the points that received the same cluster label in every iteration so far. Formally, let $X$ be the space of data points. Elements of $X$ are windowed sequences of sensor readings of length $512$. Our encoder network $f_\theta$ and clustering model $g$ are then represented by the following functions 
\begin{gather*}
    f_\theta: X \rightarrow Z\,, \\
    g: Z \rightarrow \{0, \dots, K-1\} \,,
\end{gather*}
where the latent space $Z$ is of lower dimension than $X$, $\theta$ denotes the network parameters, and $K$ is the user-defined number of clusters. Let $T$ be the total number of training iterations, and $\theta_j, 1 \leq j \leq T$ be the value of the network parameters at iteration $j$. Then, we define the full deep clustering model at iteration $j$ as
\begin{gather*}
    C_j \coloneqq g \circ f_{\theta_j}\,, \\
    C_j: X \rightarrow \{0, \dots, K-1\}\,.
\end{gather*}
The encoder loss at iteration $j+1$ is then given by
\begin{equation*}
    \mathcal{L}_j = \sum\nolimits_{M_j} \mathit{CE}(h(f_{\theta}(x_i)),C_j(x_i)) + \sum\nolimits_{S_j} \mathit{CE}(h(f_{\theta}(x_i)),C_j(x_i))\,,
\end{equation*}
where $h: Z \rightarrow (0,1)^K$ is the softmax classifier used for pseudo-label training (it is discarded after training), $\mathit{CE}$ is the categorical cross-entropy loss, 
\begin{spreadlines}{10pt}
\begin{gather*}
    M_j \coloneqq \{x \in X | C_j(x) = C_{j-1}(x)\}\,, \\
    S_j \coloneqq \{x \in X | \forall 1 \leq k < j, C_j(x) = C_{k}(x)\}\,.
\end{gather*}
\end{spreadlines}
Here, $S_j$ and $M_j$ correspond to $SemiMask$ and $Mask$ in Algorithm~\ref{alg:method}, respectively, and allow for our graded label filtering. Note that $S_j \subset M_j$.

Existing deep HAR clustering models all require an autoencoder to generate feature vectors for clustering \citep{sheng2020unsupervised,ma2021unsupervised,mcconville2021n2d}. This effectively doubles the time and space requirements, as a decoder must be trained in conjunction with the encoder. Autoencoders can also present some problems for clustering, as they learn to reconstruct every detail of the input, including irrelevant features and noise. This has been well-documented in the case of image clustering, and becomes a greater problem the larger the input is; see \citep{mrabah2020deep} and the references therein for a full discussion. Our method, in contrast, uses a single streamlined loss which does not require a decoder, and thus can scale better to richer datasets with more sensors, both in terms of computational costs and accuracy. This is supported by the results from Section~\ref{sec:results}.

Before clustering the latent space, we apply UMAP (uniform manifold approximation \citep{mcinnes2018umap}), as a second round of dimensionality reduction, reducing the latent dimension from $32$ to $2$, and cluster with a hidden Markov model (HMM) to capture temporal consistency. As with previous HAR clustering models, the number of clusters is set manually to be equal to the number of classes in the dataset.
 The label filtering method described above identifies, each time it is run, a subset of confident labels. It can thus be repeated a number of times. If, in any of the repetitions, a data point received a confident label, then the output of our model for that data point is its most recent confident label. Otherwise, the output of our model is the label from the final repetition. We iterate until the set of points that have ever received a confident label stops increasing.
 
 A final feature of our model is that, for the five smaller datasets, we reduce the step size from 100 (as is standard) to 5, to increase the number of data points. This reduction provides 20 times more data, which improves the training of the encoder, cf.\ Table \ref{tab:ablation-studies}. However HHAR, having by far the most data points, but comparatively few sensors (12, compared to, e.g., 51 for PAMAP or 117 for REALDISP), does not require additional data, so we keep the original step size of 100. The entire method is described in Algorithm \ref{alg:method}.

To evaluate the four settings discussed in Section \ref{subsec:ambiguous-eval} and presented in Table \ref{tab:baseline-results}, we use a pared-down version of our model. This pared-down version differs from the  above in that it removes UMAP and label filtering, and does not decrease the step size. That is, it simply encodes the data using the same convolutional architecture as the main model, with a step size of $100$, clusters the encodings using a HMM, and then performs pseudo-label training.

\begin{table}[]
\begin{minipage}{.55\linewidth}
\caption{Comparison with \citep{mcconville2021n2d} and various supervised methods. The supervised methods are  \protect\citep{dua2021multi} on PAMAP, UCI-Sm,  and WISDM-v1, \protect\citep{aljarrah2021human} on REALDISP, and \protect\citep{qin2020imaging} on HHAR.\vspace*{-0ex}}
\label{tab:nd2-supervised-comp}
\centering
 \resizebox{\linewidth}{!}{\begin{tabular}{*{5}{|c}|}
\hline
 & & ours & n2d & supervised \\\hline\hline
	\multirow{4}{*}{PAMAP}  & ACC & \textbf{86.3} & 86.0 & 95.3 \\
\cline{2-5}
	 & NMI & \textbf{88.4} & 85.4 & - \\
\cline{2-5}
	 & ARI & \textbf{81.4} & 78.3 & - \\
\cline{2-5}
	 & F1 & 80.1 & \textbf{83.7} & 95.2 \\
\hline
	\multirow{4}{*}{UCI-Smartphone}  & ACC & \textbf{65.9} & 52.1 & 96.2 \\
\cline{2-5}
	 & NMI & \textbf{68.6} & 55.8 & - \\
\cline{2-5}
	 & ARI & \textbf{56.3} & 38.7 & - \\
\cline{2-5}
	 & F1 & \textbf{64.4} & 48.5 & 96.2 \\
\hline
	\multirow{4}{*}{WISDM-v1}  & ACC & \textbf{75.3} & 70.5 & 97.2 \\
\cline{2-5}
	 & NMI & \textbf{76.0} & 69.1 & - \\
\cline{2-5}
	 & ARI & \textbf{65.5} & 58.1 & - \\
\cline{2-5}
	 & F1 & \textbf{68.9} & 63.4 & 97.2 \\
\hline
	\multirow{4}{*}{WISDM-watch}  & ACC & \textbf{91.7} & 84.8 & - \\
\cline{2-5}
	 & NMI & \textbf{93.8} & 88.9 & - \\
\cline{2-5}
	 & ARI & \textbf{88.1} & 79.0 & - \\
\cline{2-5}
	 & F1 & \textbf{92.4} & 84.5 & - \\
\hline
	\multirow{4}{*}{REALDISP}  & ACC & \textbf{91.0} & 80.3 & 99.8 \\
\cline{2-5}
	 & NMI & \textbf{95.2} & 90.4 & - \\
\cline{2-5}
	 & ARI & \textbf{88.9} & 79.0 & - \\
\cline{2-5}
	 & F1 & \textbf{88.0} & 72.5 & 99.8 \\
\hline
	\multirow{4}{*}{HHAR}  & ACC & \textbf{62.3} & 59.7 & 96.6 \\
\cline{2-5}
	 & NMI & \textbf{67.9} & 60.8 & - \\
\cline{2-5}
	 & ARI & \textbf{50.5} & 46.0 & - \\
\cline{2-5}
	 & F1 & 59.0 & \textbf{59.3} & 96.6 \\
\hline
	\multirow{4}{*}{UCI-full-feats}  & ACC & \textbf{65.5} & 64.9 & - \\
\cline{2-5}
	 & NMI & 59.3 & \textbf{67.0} & - \\
\cline{2-5}
	 & ARI & 46.3 & \textbf{55.1} & - \\
\cline{2-5}
	 & F1 & 64.5 & \textbf{70.4} & - \\
\hline
\end{tabular}} 
\end{minipage}
    \hspace{1em}
\begin{minipage}{.4\linewidth}
 \caption{Comparison of our method with that of \citep{sheng2020unsupervised}. Ours performs better on both datasets and all metrics, with a more significant difference  on the more complex dataset, REALDISP.\vspace*{-0ex}}
\label{tab:sheng-comp}
\centering
\begin{tabular}{*{4}{|c}|}
\hline
 & & ours & S20 \\\hline\hline
	\multirow{3}{*}{PAMAP}  & ACC & \textbf{86.3} & 85.4 \\
\cline{2-4}
	 & NMI & \textbf{88.4} & 87.3 \\
\cline{2-4}
	 & ARI & \textbf{81.4} & 80.2 \\
\hline
	\multirow{3}{*}{REALDISP}  & ACC & \textbf{91.0} & 68.1 \\
\cline{2-4}
	 & NMI & \textbf{95.2} & 60.5 \\
\cline{2-4}
	 & ARI & \textbf{88.9} & 80.4 \\
\hline
\end{tabular}\vspace*{2ex}
\caption{Comparison of our method with that of \cite{ma2021unsupervised} on the HHAR dataset. We achieve a significantly higher NMI but a lower F1.\vspace*{0ex}} \label{tab:ma-comp}

        \begin{tabular}{*{3}{|c}|}
        \hline
         & \multicolumn{2}{c|}{HHAR} \\\hline
         & F1 & NMI \\\hline\hline
        	ours & \textbf{67.9} & 59.0 \\
        \hline
        	M21 & 55.0 & \textbf{65.9} \\
        \hline
        \end{tabular}     
    \end{minipage} 
\end{table}

\section{Experimental Evaluation} \label{sec:results}
\paragraph{Metrics.}
We propose and use four metrics: clustering accuracy (ACC), adjusted Rand index (ARI), normalized mutual information (NMI), and macro-F1 (F1). After aligning predicted labels to the ground-truth labels via the alignment that maximizes accuracy, ACC and F1 are computed as in the supervised setting. ARI and NMI are computed by:
\begin{gather*}
    ARI = \frac{ \left. \sum_{ij} \binom{n_{ij}}{2} - \left[\sum_i \binom{n_i}{2} \sum_j \binom{n_j}{2}\right] \right/ \binom{n}{2} }{ \left. \frac{1}{2} \left[\sum_i \binom{n_i}{2} + \sum_j \binom{n_j}{2}\right] - \left[\sum_i \binom{n_i}{2} \sum_j \binom{n_j}{2}\right] \right/ \binom{n}{2} } \\
    NMI = \frac{ 2 \sum_i \sum_j n_{ij} \log \frac{n n_{ij}}{n_i\dot n_j}}{\sum_i n_i \log \frac{n_i}{n} + \sum_j n_j \log \frac{n_j}{n}}\,,
\end{gather*}
where $n_j$ is the number of data points in ground truth class $j$, as indicated by the labels in the dataset, $n_i$ is the number of data points cluster $i$, $n_{ij}$ is the number of data points in ground-truth class $j$ and cluster $i$, and $n$ is total number of data points.

 We choose this set of metrics, because it covers three interpretations of clustering. ARI measures performance with respect to the standard interpretation of finding a partition that maximizes intra-cluster similarity and minimizes inter-cluster similarity. Calling points with the same ground-truth label similar, and those with different labels dissimilar, ARI measures what fraction of the time similar pairs are placed together and different pairs placed separately.
NMI is based on the interpretation of clustering as compression. An accurate clustering should be able to encode the maximum amount of useful information about a given data point by specifying its cluster assignment, i.e., the cluster labels act as a compression code, replacing the data for each point with a single integer from $0, \dots, K-1$, where $K$ is the number of clusters. The ground-truth labels are thought of as the ideal compression code, and we then measure the information-theoretic distance between it and the compression code produced by the clustering model being evaluated.
ACC and F1 treat clustering as unsupervised classification. They are useful because they enable a direct comparison to supervised classifiers, as in Table \ref{tab:nd2-supervised-comp}. 
\begin{table}[h]
\caption{Ablation studies on the main components of our model.\vspace*{1ex}}
\label{tab:ablation-studies}
\centering
\resizebox{0.95\linewidth}{!}{\begin{tabular}{*{8}{|c}|}
\hline
 & & ours & no UMAP & no label-filter & GMM & step 100 & net. dim. \\
 \hline\hline
	\multirow{4}{*}{PAMAP}  & ACC & \textbf{86.3} & 56.0 & 73.3 & 78.7 & 81.6  & 65.06 \\
\cline{2-8}
	 & NMI & \textbf{88.4} & 52.9 & 76.8 & 81.0 & 81.5 & 64.32\\
\cline{2-8}
	 & ARI & \textbf{81.4} & 39.8 & 62.72 & 69.7 & 72.4 & 50.11 \\
\cline{2-8}
	 & F1 & \textbf{80.1} & 52.9 & 71.4 & 76.7 & 78.4 & 63.69\\
\hline
\multirow{4}{*}{UCI-Sm}  & ACC & \textbf{65.9} & 49.8 & 60.8 & 58.7 & 62.5 & 55.63 \\
\cline{2-8}
	 & NMI & \textbf{68.6} & 52.7 & 63.8 & 62.4 & 67.2 & 56.38\\
\cline{2-8}
	 & ARI & \textbf{56.3} & 36.9 & 49.3 & 47.1 & 52.1 & 41.71\\
\cline{2-8}
	 & F1 & \textbf{64.4} & 44.8 & 58.9 & 58.3 & 60.6 & 54.52\\
\hline
	\multirow{4}{*}{WISDM-v1}  & ACC & \textbf{75.3} & 68.8 & 68.6 & 71.5 & 69.9 & 65.72\\
\cline{2-8}
	 & NMI & \textbf{76.0} & 63.2 & 70.2 & 74.6 & 69.2 & 61.2\\
\cline{2-8}
	 & ARI & \textbf{65.5} & 55.1 & 56.0 & 60.9 & 55.9 & 51.15\\
\cline{2-8}
	 & F1 & \textbf{68.9} & 58.6 & 60.4 & 63.7 & 64.9 & 57.29\\
\hline
	\multirow{4}{*}{WISDM watch}  & ACC & \textbf{91.7} & 69.1 & 87.8 & 89.2 & 88.8 & 62.73\\
\cline{2-8}
	 & NMI & \textbf{93.8} & 80.8 & 90.5 & 92.1 & 92.6 & 70.31\\
\cline{2-8}
	 & ARI & \textbf{88.1} & 63.6 & 81.8 & 85.0 & 85.5 & 51.58\\
\cline{2-8}
	 & F1 & \textbf{92.4} & 67.0 & 88.0 & 90.4 & 89.5 & 61.9 \\
\hline
	\multirow{4}{*}{REALDISP}  & ACC & \textbf{91.0} & 82.4 & 82.7 & 87.7 & 76.8 & 51.07\\
\cline{2-8}
	 & NMI & \textbf{95.2} & 91.0 & 92.1 & 93.0 & 91.5 & 68.73\\
\cline{2-8}
	 & ARI & \textbf{88.9} & 82.7 & 79.9 & 84.5 & 79.1 & 46.53\\
\cline{2-8}
	 & F1 & \textbf{88.0} & 76.0 & 78.8 & 84.6 & 81.2 & 49.3\\
\hline
	\multirow{4}{*}{HHAR}  & ACC & \textbf{62.3} & 57.9 & 58.5 & 61.2 & -  & 64.84\\
\cline{2-8}
	 & NMI & \textbf{67.9} & 58.4 & 61.0 & 66.7 & - & 65.32\\
\cline{2-8}
	 & ARI & \textbf{50.5} & 45.8 & 45.4 & 50.2 & - & 52.3\\
\cline{2-8}
	 & F1 & \textbf{59.0} & 56.4 & 54.9 & 58.6 & - & 63.68\\
\hline
\end{tabular}}
\end{table}

 \paragraph{Network Architecture and Training Parameters.} \label{subsec:arch-and-hyperparams}
Our encoder network contains four convolutional layers, with batchnorm and max-pooling of size $2$ after each. The filter sizes and strides for the convolutional layers are $(50,2)$, $(40,2)$, $(7,1)$, $(4,1)$, and the number of filters per layer are $4,8,16,32$. In every layer, the convolutional filters are 1D with weight sharing across sensor channels. After the convolutional layers, all channels are combined with a fully connected layer (with input size  $32$ $\times $ the number of sensor channels, and output size $32$). Weights are updated by Adam \citep{kingma2014adam} with learning rate 1e-3, $\beta_1 = 0.9$, $\beta_2 = 0.999$ and weight decay of $0$. The latent space has dimension $32$. UMAP uses two components, minimum distance $0$ and n\_neighbours $60$. Clustering is performed by a hidden Markov model (HMM), whose emission probabilities are Gaussian distributions with no restrictions on the covariance matrices. The transition probabilities are set to $1-p$ on the diagonal and $\tfrac{p}{K-1}$ on the off-diagonal, where $p$ is the fraction of time points in the dataset that are followed by the same action. The window size is $512$ for all datasets. For information on the step size, see the discussion in Section \ref{sec:method}.
The softmax classifier used for pseudo-label training is a multi-layer perceptron (MLP) with a single hidden layer of $250$ units. After the encoder has been pseudo-label trained, we cluster again. We alternate ten times between clustering and pseudo-label training the encoder on the cluster labels, training for five epochs each iteration.

\paragraph{Comparisons with Prior Work.}
Tables \ref{tab:nd2-supervised-comp}, \ref{tab:sheng-comp} and \ref{tab:ma-comp} compare, respectively, the performance of our model to that of \cite{mcconville2021n2d}, \cite{sheng2020unsupervised} and \cite{ma2021unsupervised}. All evaluations are  in the pointwise subject-dependent setting. For the latter two, we are restricted to only comparing on certain datasets and metrics, as the authors do not report on all the datasets and metrics that we do, and they do not release their code. We attempted to reimplement their methods using the details  in the respective papers, but were unable to, and where we contacted the authors to ask for access to their code, we received no response. The reasons we test on these six datasets are given in Section \ref{subsec:datasets}. Table \ref{tab:nd2-supervised-comp} also compares to recent supervised models, to give an indication of the gap between them and clustering models. The supervised models used are shown in the table caption. 

We outperform \citep{mcconville2021n2d} on almost all datasets and metrics. The most significant difference is on the most complex dataset, REALDISP (see Table \ref{tab:dataset-info}). This supports that our method is better able to scale to complex datasets because it avoids the need for an autoencoder. For comparison with \citep{mcconville2021n2d}, we report on both UCI-Sm, which contains the raw sensor signal, and on UCI-feat, which contains statistical features for each window. (Both datasets and their details are available on the UCI repository.) The figures reported by \cite{mcconville2021n2d} on UCI-feat are significantly higher than we obtained by running their code, $80.1$ and $68.3$ for ACC and NMI, respectively. This difference could be due to their reported results using a larger number of clusters than the ground truth, which tends to increase ACC significantly while keeping NMI similar. For fair comparison, we fixed the number of clusters to the ground truth for all methods.

We outperform \citep{sheng2020unsupervised} on both PAMAP and REALDISP. The margin is larger for REALDISP, again suggesting that our approach is better able to leverage the more complex information in REALDISP's 117 channels, because it does not use an autoencoder. We outperform \citep{ma2021unsupervised} on NMI but not F1. Partly, this could be due to their reporting micro-F1 (not specified but implied), which tends to be higher. Our micro-F1 score on HHAR is closer at $62.30$. This difference between metrics also highlights the value of reporting multiple metrics to capture different aspects of performance.

\paragraph{Ablation Studies.}
Table \ref{tab:ablation-studies} shows the results of removing each of the components of our model. 
UMAP gives an improved performance across all metrics and datasets. This is consistent with its use in image clustering models \citep{allaoui2020considerably}. We use two different ablation settings, one removes all dimension reduction, the other replaces UMAP with and MLP of hidden size 256, and output size 2. The second setting also shows a general drop of performance, which is larger on datasets with more sensors. Results on HHAR are comparable to using UMAP, likely because, as well as only 12 sensor channels, HHAR has the most data with which the additional network can train. Results on all other datasets are worse.
PAMAP and, especially, REALDISP, show a significant drop. This suggests that, for simple datasets, more of the improvement with UMAP is due to it reducing dimension, rather than exactly how it reduces dimension, but for complex datasets with more information to fit into the reduced dimensions, the manner of dimension reduction is more important. 
Label filtering, the technique of removing likely-incorrect labels from pseudo-label training, so that the less noisy filtered labels can facilitate better training of the encoder, has been shown to improve clustering performance in prior works 
\citep{mrabah2020deep,mahon2021selective}. Here, our novel method of label-filtering is also effective, significantly increasing all metrics across all datasets.
Clustering with a HMM instead of a GMM markedly improves performance on PAMAP, UCI-Sm, and WISDM-v1. On WISDM-watch and REALDISP, where the metrics are already high, the improvement, though still significant, is less substantial. This suggests that, as the sensor data become richer, the further benefit of temporal information offers less improvement. (The GMM has full covariance matrices, convergence threshold of $1e-3$, maximum EM steps of $100$, is initialized with k-means, and takes the best clustering from five initializations.) 
The smaller step size, which produces more data points to train the encoder, is effective at improving performance on all datasets. It is most effective on the two datasets with the most sensor channels, PAMAP and REALDISP, as they require larger networks, and hence more training data.

\paragraph{Computational Costs.}
The computational costs are important to measure for HAR models, as use-cases are often on constrained hardware, such as phones and watches. Table~\ref{tab:run-time} shows the execution time (in sec) of our model, for each of the three phases of training: pseudo-label training the CNN, UMAP, and clustering. All runs use a single 12GB NVIDIA TITAN GPU. As expected, training and clustering are slower for datasets with more channels and more classes. UMAP is unaffected by either. At the time of writing, UMAP is only supported on CPU. However, a GPU implementation is expected soon, which will give a substantial speed-up. The run time is also affected by the number of iterations until convergence, as described in Section \ref{sec:method}, with PAMAP and UCI-Sm taking the most iterations.

\begin{table}[t]
    \caption{Run time, in seconds, of our model, broken down by the three phases of training. The training time is the time taken to pseudo-label train the encoder CNN.\vspace*{-0ex}} \label{tab:run-time}
    \centering
    \small
    \begin{tabular}{|r|p{.9cm}|p{.9cm}|p{.9cm}|p{.9cm}|p{1.1cm}|}
                        \hline
                        & Train         & UMAP      & Cluster       & Total         & Per Point                \\
                        \hline \hline
         PAMAP          & 13616         & 81986     & 10056         & 109523        & 5.70e-2                       \\
                        \hline
         UCI-Sm            & 4750          & 700       & 1700          & 7236          & 1.00e-1                       \\
                        \hline
         WISDM-v1       & 1896          & 18252     & 639          & 20787         & 1.90e-2                       \\
                        \hline
         WISDM-watch    & 22790         & 130580    & 40490         & 194843        & 5.36e-2                       \\
                        \hline
         REALDISP       & 3303         & 12858     & 14892         & 31053        & 4.9e-1                        \\
                        \hline
         HHAR       & 25386         & 16983     & 1137         & 43506        & 3.9e-3                        \\
                        \hline
    \end{tabular}
    \end{table}
\section{Conclusion} \label{sec:conlusion}
In this paper, we articulated and discussed the shortcomings of current evaluation procedures for HAR clustering models. We noted a lack of consistency in previous works' reporting of results, and conducted experiments to show that a common ambiguity in evaluation, namely subject dependence, can significantly alter the results. We then discussed superior evaluation alternatives. Additionally, we introduced a new deep clustering model for HAR. Tested on six public datasets, under our proposed settings, it performs better than or on par with existing works, while also being more scalable and efficient by avoiding the need for an autoencoder. 
This paper can serve as a guide for future efforts in deep clustering of human activities, by articulating the requirements for comprehensive evaluation, and by detailing a number of effective techniques with respect to our own model.

{\small 
\bibliography{bibliography}

\begin{thebibliography}{47}
\providecommand{\natexlab}[1]{#1}
\providecommand{\url}[1]{\texttt{#1}}
\expandafter\ifx\csname urlstyle\endcsname\relax
  \providecommand{\doi}[1]{doi: #1}\else
  \providecommand{\doi}{doi: \begingroup \urlstyle{rm}\Url}\fi

\bibitem[Aljarrah and Ali(2021)]{aljarrah2021human}
A.~Aljarrah and A.~Ali.
\newblock Human activity recognition by deep convolution neural networks and
  principal component analysis.
\newblock \emph{Further Advances in Internet of Things in Biomedical and Cyber
  Physical Systems}, 2021.

\bibitem[Allaoui et~al.(2020)Allaoui, Kherfi, and
  Cheriet]{allaoui2020considerably}
M.~Allaoui, M.~L. Kherfi, and A.~Cheriet.
\newblock Considerably improving clustering algorithms using umap
  dimensionality reduction technique: {A} comparative study.
\newblock In \emph{International Conference on Image and Signal Processing},
  pages 317--325. Springer, 2020.

\bibitem[Alsheikh et~al.(2016)Alsheikh, Selim, Niyato, Doyle, Lin, and
  Tan]{alsheikh2016deep}
M.~A. Alsheikh, A.~Selim, D.~Niyato, L.~Doyle, S.~Lin, and H.~P. Tan.
\newblock Deep activity recognition models with triaxial accelerometers.
\newblock In \emph{Workshops at AAAI}, 2016.

\bibitem[Altun et~al.(2010)Altun, Barshan, and
  Tun{\c{c}}el]{altun2010comparative}
K.~Altun, B.~Barshan, and O.~Tun{\c{c}}el.
\newblock Comparative study on classifying human activities with miniature
  inertial and magnetic sensors.
\newblock \emph{Pattern Recognition}, 43\penalty0 (10), 2010.

\bibitem[Anguita et~al.(2013)Anguita, Ghio, Oneto, Parra, and
  Reyes-Ortiz]{anguita2013public}
D.~Anguita, A.~Ghio, L.~Oneto, X.~Parra, and J.~Reyes-Ortiz.
\newblock A public domain dataset for human activity recognition using
  smartphones.
\newblock In \emph{ESANN}, volume~3, 2013.

\bibitem[Ba{\~n}os et~al.(2012)Ba{\~n}os, Damas, Pomares, Rojas, T{\'o}th, and
  Amft]{banos2012benchmark}
O.~Ba{\~n}os, M.~Damas, H.~Pomares, I.~Rojas, M.~T{\'o}th, and O.~Amft.
\newblock A benchmark dataset to evaluate sensor displacement in activity
  recognition.
\newblock In \emph{Proc.\ UbiComp}, 2012.

\bibitem[Caron et~al.(2018)Caron, Bojanowski, Joulin, and Douze]{caron2018deep}
M.~Caron, P.~Bojanowski, A.~Joulin, and M.~Douze.
\newblock Deep clustering for unsupervised learning of visual features.
\newblock In \emph{Proc.\ ECCV}, 2018.

\bibitem[Chen et~al.(2021)Chen, Zhang, Yao, Guo, Yu, and Liu]{chen2021deep}
K.~Chen, D.~Zhang, L.~Yao, B.~Guo, Z.~Yu, and Y.~Liu.
\newblock Deep learning for sensor-based human activity recognition: Overview,
  challenges, and opportunities.
\newblock \emph{ACM CSUR}, 2021.

\bibitem[Chen and Xue(2015)]{chen2015deep}
Y.~Chen and Y.~Xue.
\newblock A deep learning approach to human activity recognition based on
  single accelerometer.
\newblock In \emph{Proc.\ SMC}, 2015.

\bibitem[Dua et~al.(2021)Dua, Shiva~Nand, and Vijay~Bhaskar]{dua2021multi}
N.~Dua, S.~Shiva~Nand, and S.~Vijay~Bhaskar.
\newblock Multi-input {CNN-GRU} based human activity recognition using wearable
  sensors.
\newblock \emph{Computing}, 2021.

\bibitem[Ferrari et~al.(2019)Ferrari, Micucci, Mobilio, and
  Napoletano]{ferrari2019hand}
A.~Ferrari, D.~Micucci, M.~Mobilio, and P.~Napoletano.
\newblock Hand-crafted features vs residual networks for human activities
  recognition using accelerometer.
\newblock In \emph{Proc.\ ISCT}, 2019.

\bibitem[Hammerla et~al.(2016)Hammerla, Halloran, and
  Pl{\"o}tz]{hammerla2016deep}
N.~Hammerla, S.~Halloran, and T.~Pl{\"o}tz.
\newblock Deep, convolutional, and recurrent models for human activity
  recognition using wearables.
\newblock \emph{arXiv:1604.08880}, 2016.

\bibitem[He et~al.(2017)He, Tan, and Huang]{he2017unsupervised}
H.~He, Y.~Tan, and J.~Huang.
\newblock Unsupervised classification of smartphone activities signals using
  wavelet packet transform and half-cosine fuzzy clustering.
\newblock In \emph{Proc.\ FUZZ-IEEE}, 2017.

\bibitem[He et~al.(2018)He, Tan, and Zhang]{he2018wavelet}
H.~He, Y.~Tan, and W.~Zhang.
\newblock A wavelet tensor fuzzy clustering scheme for multi-sensor human
  activity recognition.
\newblock \emph{Engineering Applications of Artificial Intelligence}, 2018.

\bibitem[He and Jin(2009)]{he2009activity}
Z.~He and L.~Jin.
\newblock Activity recognition from acceleration data based on discrete consine
  transform and svm.
\newblock In \emph{Proc.\ SMC}, 2009.

\bibitem[Hongeng et~al.(2000)Hongeng, Bremond, and
  Nevatia]{hongeng2000representation}
S.~Hongeng, F.~Bremond, and R.~Nevatia.
\newblock Representation and optimal recognition of human activities.
\newblock In \emph{Proc.\ CVPR}, volume~1, 2000.

\bibitem[Inoue et~al.(2018)Inoue, Inoue, and Nishida]{inoue2018deep}
M.~Inoue, S.~Inoue, and T.~Nishida.
\newblock Deep recurrent neural network for mobile human activity recognition
  with high throughput.
\newblock \emph{AROB}, 23\penalty0 (2), 2018.

\bibitem[Jiang and Yin(2015)]{jiang2015human}
W.~Jiang and Z.~Yin.
\newblock Human activity recognition using wearable sensors by deep
  convolutional neural networks.
\newblock In \emph{Proc.\ ACM Multimedia}, 2015.

\bibitem[Khan et~al.(2010)Khan, Lee, Lee, and Kim]{khan2010human}
A.~M. Khan, Y.~K. Lee, S.~Y. Lee, and T.~S. Kim.
\newblock Human activity recognition via an accelerometer-enabled-smartphone
  using kernel discriminant analysis.
\newblock In \emph{Proc.\ ICFIT}, 2010.

\bibitem[Kingma and Ba(2014)]{kingma2014adam}
D.~Kingma and J.~Ba.
\newblock Adam: A method for stochastic optimization.
\newblock \emph{arXiv preprint arXiv:1412.6980}, 2014.

\bibitem[Kwapisz et~al.(2011)Kwapisz, Weiss, and Moore]{kwapisz2011activity}
J.~Kwapisz, G.~Weiss, and S.~Moore.
\newblock Activity recognition using cell phone accelerometers.
\newblock \emph{ACM SigKDD Explorations Newsletter}, 12\penalty0 (2), 2011.

\bibitem[Kwon et~al.(2014)Kwon, Kang, and Bae]{kwon2014unsupervised}
Y.~Kwon, K.~Kang, and C.~Bae.
\newblock Unsupervised learning for human activity recognition using smartphone
  sensors.
\newblock \emph{Expert Systems with Applications}, 41\penalty0 (14), 2014.

\bibitem[Li and Dustdar(2011)]{li2011incorporating}
F.~Li and S.~Dustdar.
\newblock Incorporating unsupervised learning in activity recognition.
\newblock In \emph{Workshops at AAAI}, 2011.

\bibitem[Li et~al.(2014)Li, Shi, Ding, and Liu]{li2014unsupervised}
Y.~Li, D.~Shi, B.~Ding, and D.~Liu.
\newblock Unsupervised feature learning for human activity recognition using
  smartphone sensors.
\newblock In \emph{Proc.\ MIKE}. 2014.

\bibitem[Lu et~al.(2017)Lu, Wei, Liu, Zhong, Sun, and Liu]{lu2017towards}
Y.~Lu, Y.~Wei, L.~Liu, J.~Zhong, L.~Sun, and Y.~Liu.
\newblock Towards unsupervised physical activity recognition using smartphone
  accelerometers.
\newblock \emph{Multimedia Tools and Applications}, 2017.

\bibitem[M.~Zhang(2011)]{zhang2011feature}
A.~Sawchuk M.~Zhang.
\newblock A feature selection-based framework for human activity recognition
  using wearable multimodal sensors.
\newblock In \emph{BodyNets}, 2011.

\bibitem[Ma et~al.(2021)Ma, Zhang, Li, and Lu]{ma2021unsupervised}
H.~Ma, Z.~Zhang, W.~Li, and S.~Lu.
\newblock Unsupervised human activity representation learning with multi-task
  deep clustering.
\newblock 2021.

\bibitem[Machado et~al.(2015)Machado, Gomes, Gamboa, Paix{\~a}o, and
  Costa]{machado2015human}
I.~Machado, A.~Gomes, H.~Gamboa, V.~Paix{\~a}o, and R.~Costa.
\newblock Human activity data discovery from triaxial accelerometer sensor:
  Non-supervised learning sensitivity to feature extraction parametrization.
\newblock \emph{Information Processing \& Management}, 51\penalty0 (2), 2015.

\bibitem[Mahon and Lukasiewicz(2021)]{mahon2021selective}
L.~Mahon and T.~Lukasiewicz.
\newblock Selective pseudo-label clustering.
\newblock In \emph{Proc.\ KI}, 2021.

\bibitem[Malekzadeh et~al.(2018)Malekzadeh, Clegg, Cavallaro, and
  Haddadi]{malekzadeh2018protecting}
M.~Malekzadeh, R.~Clegg, A.~Cavallaro, and H.~Haddadi.
\newblock Protecting sensory data against sensitive inferences.
\newblock In \emph{Workshops at EuroSys}, 2018.

\bibitem[McConnell et~al.(2018)McConnell, Turakhia, Harrington, King, and
  Ashley]{mcconnell2018mobile}
M.~McConnell, M.~Turakhia, R.~Harrington, A.~King, and E.~Ashley.
\newblock Mobile health advances in physical activity, fitness, and atrial
  fibrillation: moving hearts.
\newblock \emph{JACC}, 2018.

\bibitem[McConville et~al.(2021)McConville, Santos-Rodriguez, Piechocki, and
  Craddock]{mcconville2021n2d}
R.~McConville, R.~Santos-Rodriguez, R.~J. Piechocki, and I.~Craddock.
\newblock N2d:(not too) deep clustering via clustering the local manifold of an
  autoencoded embedding.
\newblock In \emph{2020 25th International Conference on Pattern Recognition
  (ICPR)}, pages 5145--5152. IEEE, 2021.

\bibitem[McInnes et~al.(2018)McInnes, Healy, and Melville]{mcinnes2018umap}
L.~McInnes, J.~Healy, and J.~Melville.
\newblock Umap: Uniform manifold approximation and projection for dimension
  reduction.
\newblock \emph{arXiv:1802.03426}, 2018.

\bibitem[Mejia-Ricart et~al.(2017)Mejia-Ricart, Helling, and
  Olmsted]{mejia2017evaluate}
L.~Mejia-Ricart, P.~Helling, and A.~Olmsted.
\newblock Evaluate action primitives for human activity recognition using
  unsupervised learning approach.
\newblock In \emph{Proc.\ ICITST}, 2017.

\bibitem[Mrabah et~al.(2020)Mrabah, Khan, Ksantini, and
  Lachiri]{mrabah2020deep}
N.~Mrabah, N.~Khan, R.~Ksantini, and Z.~Lachiri.
\newblock Deep clustering with a dynamic autoencoder: From reconstruction
  towards centroids construction.
\newblock \emph{Neural Networks}, 130, 2020.

\bibitem[Qin et~al.(2020)Qin, Zhang, Meng, Qin, and Choo]{qin2020imaging}
Z.~Qin, Y.~Zhang, S.~Meng, Z.~Qin, and K.-K.~R. Choo.
\newblock Imaging and fusing time series for wearable sensor-based human
  activity recognition.
\newblock \emph{Information Fusion}, 53:\penalty0 80--87, 2020.

\bibitem[Reiss and Stricker(2012)]{reiss2012introducing}
A.~Reiss and D.~Stricker.
\newblock Introducing a new benchmarked dataset for activity monitoring.
\newblock In \emph{Proc.\ ISWC}, 2012.

\bibitem[Ronao and Cho(2015)]{ronao2015deep}
C.~A. Ronao and S.-B. Cho.
\newblock Deep convolutional neural networks for human activity recognition
  with smartphone sensors.
\newblock In \emph{Proc.\ NeurIPS}, 2015.

\bibitem[Sheng and Huber(2020)]{sheng2020unsupervised}
T.~Sheng and M.~Huber.
\newblock Unsupervised embedding learning for human activity recognition using
  wearable sensor data.
\newblock In \emph{Proc.\ FLAIRS}, 2020.

\bibitem[Singh et~al.(2017)Singh, Merdivan, Psychoula, Kropf, Hanke, Geist, and
  Holzinger]{singh2017human}
D.~Singh, E.~Merdivan, I.~Psychoula, J.~Kropf, S.~Hanke, M.~Geist, and
  A.~Holzinger.
\newblock Human activity recognition using recurrent neural networks.
\newblock In \emph{Proc.\ CD-MAKE}, 2017.

\bibitem[Stisen et~al.(2015)]{stisen2015smart}
A.~Stisen et~al.
\newblock Smart devices are different: Assessing and mitigating mobile sensing
  heterogeneities for activity recognition.
\newblock In \emph{Proc.\ SenSys}, 2015.

\bibitem[Suh et~al.(2021)Suh, Rey, and Lukowicz]{suh2021adversarial}
S.~Suh, V.~Rey, and P.~Lukowicz.
\newblock Adversarial deep feature extraction network for user independent
  human activity recognition.
\newblock \emph{arXiv preprint arXiv:2110.12163}, 2021.

\bibitem[Tapia et~al.(2004)Tapia, Intille, and Larson]{tapia2004activity}
E.~Tapia, S.~Intille, and K.~Larson.
\newblock Activity recognition in the home using simple and ubiquitous sensors.
\newblock In \emph{Proc.\ PerCom}, 2004.

\bibitem[Trabelsi et~al.(2013)Trabelsi, Mohammed, Chamroukhi, Oukhellou, and
  Amirat]{trabelsi2013unsupervised}
D.~Trabelsi, S.~Mohammed, F.~Chamroukhi, L.~Oukhellou, and Y.~Amirat.
\newblock An unsupervised approach for automatic activity recognition based on
  hidden markov model regression.
\newblock \emph{T-ASE}, 2013.

\bibitem[Wang et~al.(2019)Wang, Chen, Hao, Peng, and Hu]{wang2019deep}
J.~Wang, Y.~Chen, S.~Hao, X.~Peng, and L.~Hu.
\newblock Deep learning for sensor-based activity recognition: A survey.
\newblock \emph{Pattern Recognition Letters}, 119, 2019.

\bibitem[Weiss et~al.(2019)Weiss, Yoneda, and Hayajneh]{weiss2019smartphone}
G.~Weiss, K.~Yoneda, and T.~Hayajneh.
\newblock Smartphone and smartwatch-based biometrics using activities of daily
  living.
\newblock \emph{IEEE Access}, 7, 2019.

\bibitem[Yang et~al.(2015)Yang, Nguyen, San, Li, and
  Krishnaswamy]{yang2015deep}
J.~Yang, M.~N. Nguyen, P.~P. San, X.~L. Li, and S.~Krishnaswamy.
\newblock Deep convolutional neural networks on multichannel time series for
  human activity recognition.
\newblock In \emph{Proc.\ IJCAI}, 2015.

\end{thebibliography}
}
\end{document}